\documentclass[twocolumn, switch]{article} 

\usepackage{preprint}

\usepackage{algorithm}
\usepackage{algorithmic}
\usepackage{amsmath}
\usepackage{amsfonts}
\usepackage{mathtools}
\usepackage{multirow}

\usepackage[numbers,square,sort]{natbib}

\usepackage[utf8]{inputenc}	
\usepackage[T1]{fontenc}	
\usepackage{xcolor}		
\usepackage[colorlinks = true,
            linkcolor = purple,
            urlcolor  = blue,
            citecolor = cyan,
            anchorcolor = black]{hyperref}	
\usepackage{booktabs} 		
\usepackage{nicefrac}		
\usepackage{microtype}		
\usepackage{lineno}		
\usepackage{float}			

\usepackage{newfloat}
\DeclareFloatingEnvironment[name={Supplementary Figure}]{suppfigure}
\usepackage{sidecap}
\sidecaptionvpos{figure}{c}

\usepackage{titlesec}
\titlespacing\section{0pt}{12pt plus 3pt minus 3pt}{1pt plus 1pt minus 1pt}
\titlespacing\subsection{0pt}{10pt plus 3pt minus 3pt}{1pt plus 1pt minus 1pt}
\titlespacing\subsubsection{0pt}{8pt plus 3pt minus 3pt}{1pt plus 1pt minus 1pt}

\title{Guided and Variance-Corrected Fusion with One-shot Style Alignment for Large-Content Image Generation}
\date{}

\usepackage{authblk}

\author[1]{Shoukun Sun}
\author[1\thanks{\tt{mxian@uidaho.edu}}]{Min Xian}
\author[2]{Tiankai Yao}
\author[2]{Fei Xu}
\author[2]{Luca Capriotti}

\affil[1]{Department of Computer Science, University of Idaho}
\affil[2]{Idaho National Laboratory}

\begin{document}

\twocolumn[ 
  \begin{@twocolumnfalse} 
  
\maketitle
\begin{abstract}
Producing large images using small diffusion models is gaining increasing popularity, as the cost of training large models could be prohibitive. A common approach involves jointly generating a series of overlapped image patches and obtaining large images by merging adjacent patches. However, results from existing methods often exhibit noticeable artifacts, e.g., seams and inconsistent objects and styles. To address the issues, we proposed Guided Fusion (GF), which mitigates the negative impact from distant image regions by applying a weighted average to the overlapping regions. Moreover, we proposed Variance-Corrected Fusion (VCF), which corrects data variance at post-averaging, generating more accurate fusion for the Denoising Diffusion Probabilistic Model. Furthermore, we proposed a one-shot Style Alignment (SA), which generates a coherent style for large images by adjusting the initial input noise without adding extra computational burden. Extensive experiments demonstrated that the proposed fusion methods improved the quality of the generated image significantly. The proposed method can be widely applied as a plug-and-play module to enhance other fusion-based methods for large image generation. Code: \href{https://github.com/TitorX/GVCFDiffusion}{https://github.com/TitorX/GVCFDiffusion}
\end{abstract}
\vspace{0.35cm}

  \end{@twocolumnfalse} 
] 

\begin{figure*}[t!]
    \centering
    \includegraphics[width=0.9\linewidth]{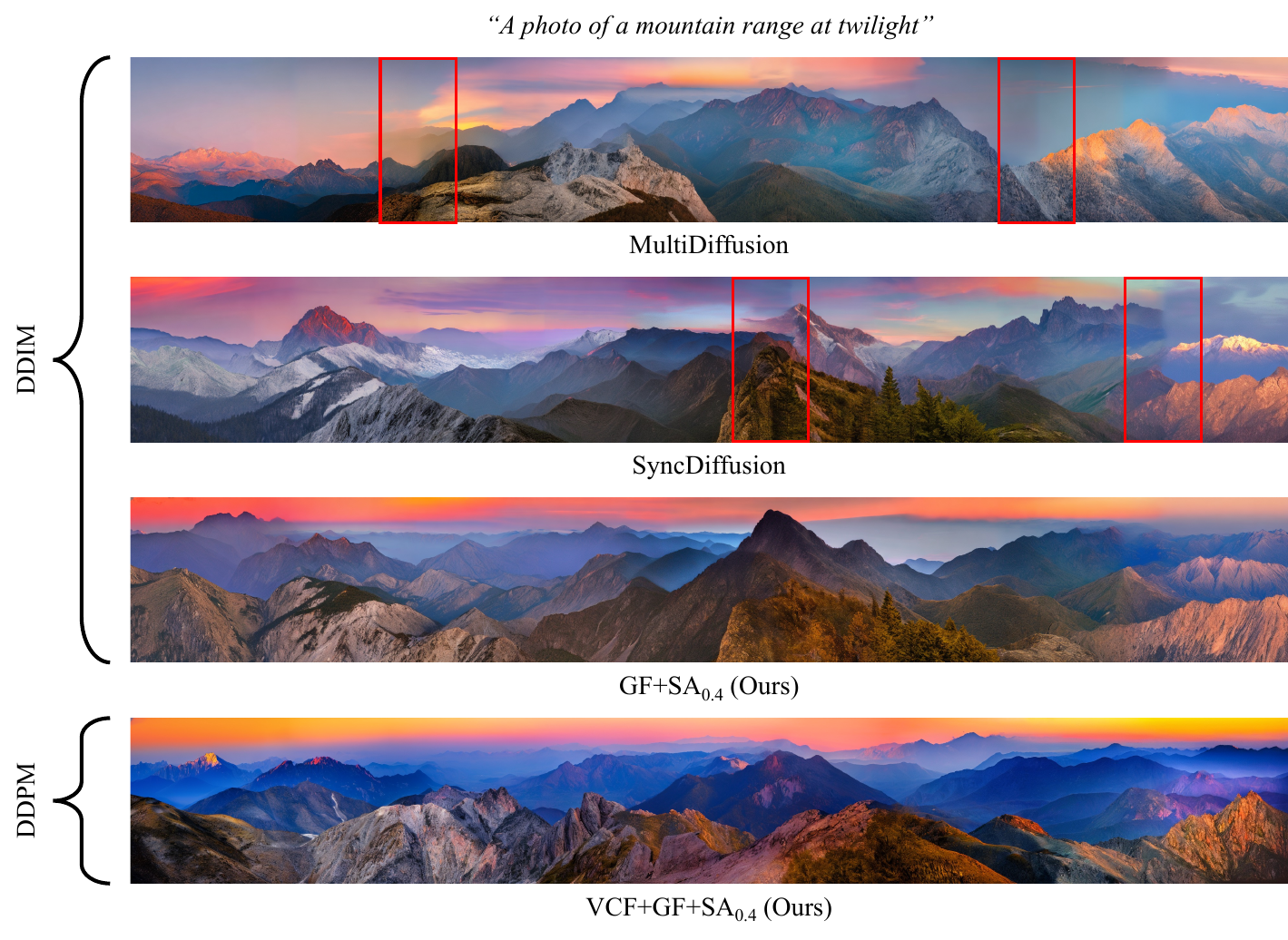}
    \caption{Comparisons of panorama images generated by MultiDiffusion \cite{bar-tal_multidiffusion_2023}, SyncDiffusion \cite{lee_syncdiffusion_2023} and our methods: Guided Fusion (GF), Variance-Corrected Fusion (VCF) and Style Alignment (SA). All images are generated with the same initial noise. The red boxes highlight the discontinuous and defective areas on the generated image.}
    \label{fig:overall}
\end{figure*}

\section{Introduction}

Recent years have witnessed remarkable advancements in text-to-image generation models, which can produce realistic and diverse images based on textual prompts. Among them, the Diffusion models, specifically the Stable Diffusion (SD) \cite{rombach_high-resolution_2022}, have emerged as one of the mainstream methods for image generation. 

There is a significant demand for producing large images. The pursuit of generating larger images involves 1) producing images with higher resolution that exhibit ultra-fine details and 2) creating images that encompass more content, such as panorama images. To differentiate between these aspects, we refer to them as High-Resolution image generation and Large-Content image generation, respectively. However, training models capable of generating large images requires a substantial investment in hardware and data. For instance, training the SD v2 model to generate $512^2$ images took over a month on 256 A100 GPUs. The core U-Net model of it comprises 865 million parameters. The larger SDXL \cite{podell_sdxl_2023} model, which can generate $1024^2$ images and contains 2.6 billion parameters, demands a longer training period.

Recent progress has been made by using pre-trained smaller models to jointly generate a series of overlapped small patches, which are then combined to form images of arbitrary sizes. A notable work is MultiDiffusion \cite{bar-tal_multidiffusion_2023}, which generates large images by averaging overlapped areas of patches at each denoising step. SyncDiffusion \cite{lee_syncdiffusion_2023} achieves more coherent large-content images by ensuring consistent styles across each small patch during the joint denoising process. However, existing methods exhibit three major drawbacks: 1) noticeable seams at overlapped areas, 2) generation of discontinuous objects, and 3) low-quality content.

Each patch derives different values in the overlapped regions at each denoising step. Resolving discrepancies by averaging to achieve uniformity values can interfere with the denoising of individual patches. This interference occurs because diffusion models, during training, assume that the whole denoising process is completed with all intermediate results undisturbed. Persistent changes to the values in certain regions can have unknown impacts on the denoising process, typically resulting in adverse effects.

We propose a method termed Guided Fusion (GF), which assigns a guidance map to each small patch to perform weighted averaging in the overlapped regions. This allows the denoising process to be dominated by patches close to the center (i.e., higher weights). Additionally, we discovered that averaging the overlapped regions while using Stochastic Differential Equation (SDE) samplers, such as Denoising Diffusion Probabilistic Model (DDPM) \cite{ho_denoising_2020}, produces highly blurred results. This occurs because the SDE samplers usually introduce a Gaussian-distributed random term during the denoising process, and averaging multiple variables sampled from Gaussian distributions results in a variance lower than expected, leading to blurred images that lack details. To address this, we introduce Variance-Corrected Fusion (VCF) to adjust the variance and generate higher-quality images. Furthermore, we observed that significant differences in the initial noise each patch uses make it more challenging to produce coherent images. Therefore, we propose a one-shot Style Alignment (SA), which aligns the initial noise with semantic interpolation to produce more style-consistent results.

The main contributions of this paper are as follows: 

\begin{itemize}
    \item Guided Fusion was proposed to utilize a guidance map for weighted averaging on overlapped areas, leading to better quality and seamless image generation.
    \item We proposed the Variance-Corrected Fusion to fix the small variance issue while averaging overlapped regions with SDE samplers. The proposed method prevents generating blurred results with SDE samplers, leading to higher-quality image generation.
    \item We proposed the one-shot Style Alignment approach that aligns the style of the initial noise only once to generate more coherent content without increasing the computational burden.
\end{itemize}

\section{Preliminaries}

The core of diffusion models (DMs) lies in the concept of a Markov process, specifically, a type of Markov chain where each step adds a controlled amount of Gaussian noise to the data. The forward diffusion process is defined as a sequence of latent variables $\{\mathbf{x}_t\}$ indexed by discrete time steps $t = 0, 1, \ldots, T$, where $\mathbf{x}_0$ represents the original data and $\mathbf{x}_T$ approximates a standard Gaussian distribution $\mathcal{N}(\mathbf{0}, \mathbf{I})$. The transition from $\mathbf{x}_{t-1}$ to $\mathbf{x}_{t}$ is modeled by a Gaussian distribution, typically formulated as:

\begin{equation}
\label{eq:forward}
q(\mathbf{x}_{t} | \mathbf{x}_{t-1}) := \mathcal{N}(\mathbf{x}_{t}; \sqrt{1 - \beta_t} \mathbf{x}_{t-1}, \beta_t \mathbf{I}).
\end{equation}

Here, the schedule of variances $\beta_t$ is designed to gradually add noise to $\mathbf{x}_t$, which can be learned by reparameterization \cite{kingma_auto-encoding_2013} or held a sequence of constants as hyperparameters \cite{rombach_high-resolution_2022, song_denoising_2020}. The choice of the $\{\beta_t\}$ is critical as it controls the rate at which the data is diffused into noise over time.

The reverse diffusion process, or called the denoising process, involves learning a model $p_\theta(\mathbf{x}_{t-1} | \mathbf{x}_t)$ that approximates the reverse of the forward process. This is done by parameterizing the Gaussian distribution with learnable parameters \(\theta\), usually expressed as

\begin{equation}
p_\theta(\mathbf{x}_{t-1} | \mathbf{x}_t) := \mathcal{N}(\mathbf{x}_{t-1}; \mathbf{\mu}_\theta(\mathbf{x}_t, t), \mathbf{\Sigma}_\theta(\mathbf{x}_t, t)),
\end{equation}

\noindent where $\mathbf{\mu}_\theta(\mathbf{x}_t, t)$ and $\mathbf{\Sigma}_\theta(\mathbf{x}_t, t)$ are learned through optimization. The objective is to minimize the difference between the true reverse distribution $q(\mathbf{x}_{t-1} | \mathbf{x}_t, \mathbf{x}_0)$ and the modeled distribution $p_\theta(\mathbf{x}_{t-1} | \mathbf{x}_t)$.

A common practice sets the schedule of $\beta_t$ as an increasing sequence of constants at the forward process. The reverse process sets $\mathbf{\Sigma}_\theta(\mathbf{x}_t, t) = \sigma_t^2 \mathbf{I}$ and let $\sigma_t^2 = \beta_t$ or $\sigma_t^2 = \frac{1 - \bar{\alpha}_{t-1}}{1-\bar{\alpha}_t} \beta_t$ \cite{ho_denoising_2020}, where $\bar{\alpha}_t = \prod_{s=1}^t\alpha_s$ and $\alpha_t := 1 - \beta_t$. Hence, we can formulate:

\begin{equation}
\label{eq:reverse}
p_\theta(\mathbf{x}_{t-1} | \mathbf{x}_t) := \mathcal{N}(\mathbf{x}_{t-1}; \mathbf{\mu}_\theta(\mathbf{x}_t, t), \sigma_t^2 \mathbf{I}).
\end{equation}

Latent Diffusion Model (LDM) \cite{rombach_high-resolution_2022} extends diffusion models by operating in a low-dimensional latent space instead of the high-dimensional pixel space. This is achieved by first encoding the data into a latent representation using a suitable encoder and then applying the diffusion process within this more compact space. This reduction in dimensionality leads to more efficient modeling and sampling as the model needs to learn and operate over fewer parameters. The Variational Autoencoders (VAEs) \cite{kingma_auto-encoding_2013} are often chosen for encoding images to latent space and decoding to pixel space.

\begin{figure}[h]
    \centering
    \includegraphics[width=0.7\linewidth]{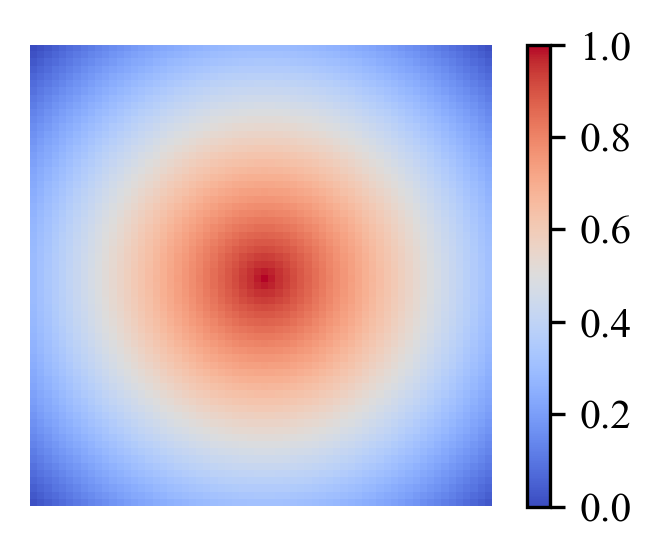}
    \caption{Guided Fusion Map.}
    \label{fig:guidance-map}
\end{figure}

\begin{figure}[h]
    \centering
    \includegraphics[width=\linewidth]{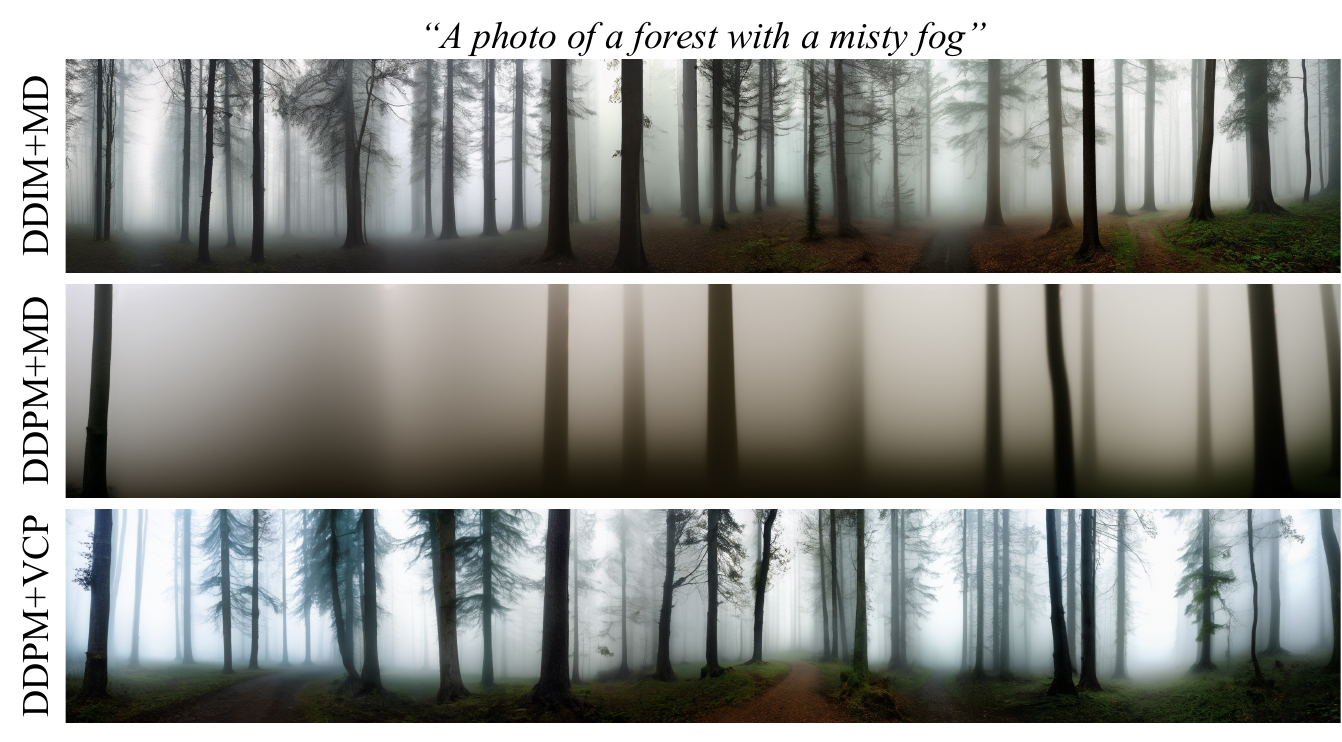}
    \caption{Images produced by direct averaging overlapped areas with DDIM and DDPM sampler, and a result from DDPM with Variance-Corrected Fusion (VCF).}
    \label{fig:avg_wrong}
\end{figure}

\begin{table*}[ht]
\centering
\begin{tabular}{c|c|ccccc}
\hline
Stride               & Fusion & FID↓           & KID↓($\times 10^{-3}$) & GIQA-QS↑       & GIQA-DS↑       & CLIP↑          \\
\hline\hline
\multirow{2}{*}{128} & MD     & 20.60          & 9.14                   & 9.311          & 9.203          & \textbf{31.65} \\
                     & GF     & \textbf{17.64} & \textbf{7.72}          & \textbf{9.324} & \textbf{9.218} & 31.59          \\
\hline
\multirow{2}{*}{256} & MD     & 17.32          & 7.21                   & 9.183          & 9.117          & \textbf{31.58} \\
                     & GF     & \textbf{15.99} & \textbf{6.68}          & \textbf{9.280} & \textbf{9.188} & 31.52          \\
\hline
\multirow{2}{*}{384} & MD     & 15.55          & 6.49                   & 9.208          & 9.136          & 31.50          \\
                     & GF     & \textbf{14.88} & \textbf{6.28}          & \textbf{9.236} & \textbf{9.159} & \textbf{31.51} \\
\hline\hline
\end{tabular}
\caption{Quantitative comparisons between MultiDiffusion (MD) \cite{bar-tal_multidiffusion_2023} and Guided Fusion (GF) with DDIM sampler using various strides. The best results within each stride group are marked in bold.}
\label{tab:gf}
\end{table*}

\section{Method}

\textbf{\textit{The nature of the joint denoising process}}. We denote a small pretrained diffusion model as a parametric model that has been optimized for a series of Markov chained Gaussian transitions $p_\theta(\mathbf{x}_0) := p_\theta(\mathbf{x}_{0:T})d\mathbf{x}_{1:T}$ at a low-dimensional space $\mathbf{x}_0 \in \mathbb{R}^n$. As the small diffusion model has never been optimized with the high-dimensional dataset, it cannot be directly used to sample larger images. The joint denoising process uses the small model to obtain large images $\mathbf{X}_0 \in \mathbb{R}^m$, where $m > n$, by fusing a series of overlapped patches after each denoising step. Since the distribution in high-dimensional space is unknown, we can only aim to sample a $\mathbf{X}_0$ for which each subview: 1) conforms to a learned distribution in the low-dimensional space so that each generated patch is realistic; 2) shares identical values in the overlapping dimensions so that can be merged to form a large sample.

\textbf{\textit{The drawbacks of averaging latent variables}}. Use a simple case as an illustration, we denote a large sample with three dimensions as $\mathbf{X}=[x^{(1)}, x^{(2)}, x^{(3)}]$ and use a two-dimensional model to jointly produce overlapped patches $\mathbf{x}^{(1)} = [x^{(1)}, x^{(2)}]$ and $\mathbf{x}^{(2)} = [x^{(2)}, x^{(3)}]$. The MultiDiffusion \cite{bar-tal_multidiffusion_2023} introduced a joint denoising process that averages values on overlapped dimensions after each denoising step, which can be described as:

\begin{equation}
\label{eq:diverge}
\begin{aligned}
    [x^{(1)}_{t-1}, x^{(21)}_{t-1}] &\sim p_\theta(\mathbf{x}_{t-1}^{(1)}|\mathbf{x}^{(1)}_{t}) \\
    [x^{(22)}_{t-1}, x^{(3)}_{t-1}] &\sim p_\theta(\mathbf{x}_{t-1}^{(2)}|\mathbf{x}^{(2)}_{t})
\end{aligned}
\end{equation}

\begin{equation}
\label{eq:avg}
    x^{(2)}_{t-1} = \frac{x^{(21)}_{t-1} + x^{(22)}_{t-1}}{2}
\end{equation}

\begin{equation}
\begin{aligned}
    \mathbf{x}^{(1)}_{t-1} &:= \mathbf{\widetilde{x}}^{(1)}_{t-1} = [x^{(1)}_{t-1}, x^{(2)}_{t-1}] \\
    \mathbf{x}^{(2)}_{t-1} &:= \mathbf{\widetilde{x}}^{(2)}_{t-1} = [x^{(2)}_{t-1}, x^{(3)}_{t-1}].
\end{aligned}
\end{equation}

As shown in Eq. \ref{eq:diverge}, the denoising steps for $\mathbf{x}^{(1)}_t$ and $\mathbf{x}^{(2)}_t$ produce diverged values $x^{(21)}_{t-1}$ and $x^{(22)}_{t-1}$ over the same dimension $x^{(2)}$. Averaging by Eq. \ref{eq:avg} solves the divergence to ensure the overlapped dimension shares the same value after each step.

As described in Eq. \ref{eq:reverse}, throughout the denoising process, for $1 < t < T$, $\mathbf{x}_{t-1}$ should be estimated by the conditional probability $p_\theta(\mathbf{x}_{t-1} | \mathbf{x}_t)$. However, during the patch averaging, the values of overlapped dimensions have been constantly modified, leading to the next $\mathbf{x}_{t-1}$ being estimated conditioned at an altered $\mathbf{\widetilde{x}}_t$. Such value altering perturbs the denoising transitions, leading to obvious seams and reduced quality.

\subsection{Mitigate Divergence among Patches with Guided Fusion}

The patch fusion using averaging could significantly alter the patch values and shift the input distribution. The diffusion model has not been fine-tuned in a training-free setting for such modified input. Hence, the model may generate degenerated results. To mitigate the input distribution shifting issue, we propose a guidance map as shown in Figure \ref{fig:guidance-map}, which linearly decreases its weight from 1 at the center to 0 at the corners to guide the weighted averaging of the overlapping regions. The weighted averaging allows the altered input to attach more to the closer patches, reducing the significant value changes.

Disrupting the denoising process of a patch in different regions may lead to varying degrees of model performance degradation. Intuitively, we consider that the closer the disturbed area is to the center, the greater the impact on the quality of the generated image. Therefore, we propose a guidance map as shown in Figure \ref{fig:guidance-map}, which linearly decreases its weight from 1 at the center to 0 at the corners to guide the weighted averaging of the overlapping regions. Following the example described by Eq. \ref{eq:avg}, the weighted average at overlapped dimension can be formulated as:

\begin{equation}
    x^{(2)}_{t-1} = \frac{w_1 x^{(21)}_{t-1} + w_2 x^{(22)}_{t-1}}{w_1 + w_2}
\end{equation}

\noindent where the corresponding locations on the guidance map determine the weights $w_1$ and $w_2$. To generalize the simple case to N overlapped patches, we formulate the weighted average for each dimension from overlapped areas as:

\begin{equation}
\label{eq:gf}
    x_{t-1} = \frac{\sum_{i}^{N} w_i x^{(i)}_{t-1}}{\sum_i^N w_i}.
\end{equation}

This method is called Guided Fusion (GF). During the joint denoising process, the value of each dimension in the overlapped area is predominantly determined by the geometrically closer patch, thereby reducing the perturbation in the denoising process for that dimension.

\begin{table*}[h]
\centering
\begin{tabular}{c|c|c|c|c|c|c}
\hline
Samplers                      & Methods                    & FID↓              & KID↓ ($\times 10^{-3}$) & GIQA-QS↑          & GIQA-DS↑          & CLIP↑             \\
\hline\hline
\multirow{5}{*}{DDIM}         & MD                         & 11.22             & 3.47                    & 8.952             & 8.870             & 31.69             \\
                              & Sync                       & 11.19             & 3.34                    & 8.965             & \underline{8.882} & \underline{31.70} \\
                              & Elastic                    & 122.52            & 53.88                   & 6.562             & 6.711             & 22.46             \\
                              & GF (Ours)                  & \underline{10.71} & \textbf{3.10}           & \textbf{8.976}    & \textbf{8.887}    & 31.67             \\
                              & MD + SA$_{0.4}$ (Ours)     & 10.82             & 3.27                    & 8.948             & 8.866             & \textbf{31.71}    \\
                              & GF + SA$_{0.4}$ (Ours)     & \textbf{10.40}    & \underline{3.12}        & \underline{8.970} & 8.877             & 31.68             \\
\hline\hline
\multirow{4}{*}{DDPM (Ours) } & VCF                        & 4.78              & 1.29                    & 8.987             & 8.970             & \underline{31.87} \\
                              & VCF + GF                   & \underline{4.35}  & \underline{1.09}        & \textbf{9.001}    & \textbf{8.981}    & 31.85             \\
                              & VCF + SA$_{0.4}$           & 4.43              & 1.19                    & 8.977             & 8.962             & \textbf{31.90}    \\
                              & VCF + GF + SA$_{0.4}$      & \textbf{4.02}     & \textbf{1.02}           & \underline{8.998} & \underline{8.976} & 31.86             \\
\hline\hline
\end{tabular}
\caption{Overall performance. The subscript of SA indicates the value of $\alpha$. Each sampler group's best and second results are marked in bold and underlined, respectively.}
\label{tab:overall}
\end{table*}

\subsection{Correcting Variance of Fused Patches with SDE Samplers}

For Ordinary Differential Equation (ODE) samplers, such as the Denoising Diffusion Implicit Model (DDIM) \cite{song_denoising_2020}, the experimental results demonstrate that although averaging fusion interferes with the denoising process, it can still produce compelling images, as shown in the first row of Figure \ref{fig:avg_wrong}. However, for scenarios requiring the use of Stochastic Differential Equation (SDE) samplers, such as DDPM \cite{ho_denoising_2020}, averaging can lead to faulty blurred results, as displayed in the second row of Figure \ref{fig:avg_wrong}. We use DDPM as an example to illustrate the reason.

For a single image patch generation using DDPM, the $t-1$ denoised image is computed by:

\begin{equation}
    \mathbf{x}_{t-1}=\frac{1}{\sqrt{\alpha_t}}\left(\mathbf{x}_t-\frac{\beta_t}{\sqrt{1-\bar{\alpha}_t}} \boldsymbol{\epsilon}_\theta\left(\mathbf{x}_t, t\right)\right)+\sigma_t \mathbf{z}
\end{equation}

\noindent where $\mathbf{z} \sim \mathcal{N}(\mathbf{0}, \mathbf{I})$. We can consider $\mathbf{x}_t$ as a known variable because the previous step has determined it, hence the: 

\begin{equation}
\label{eq:expected_variance}
    \mathbf{x}_{t-1} \sim \mathcal{N}(\mu_t, \sigma_t^2)
\end{equation}

\noindent where $\mu_t = \frac{1}{\sqrt{\alpha_t}}\left(\mathbf{x}_t-\frac{\beta_t}{\sqrt{1-\bar{\alpha}_t}} \boldsymbol{\epsilon}_\theta\left(\mathbf{x}_t, t\right)\right)$.

Continuing the example from the Eq. \ref{eq:avg} using DDPM sampler, the fused denoised dimension $x^{(2)}_{t-1} = \frac{x^{(21)}_{t-1} + x^{(22)}_{t-1}}{2}$ has:

\begin{equation}
    x^{(2)}_{t-1} \sim \mathcal{N}(\frac{\mu_{t}^{(21)} + \mu_{t}^{(22)}}{2}, \frac{\sigma_t^2}{2}).
\end{equation}

We notice that the variance becomes $\sigma_t^2/2$, smaller than the expected $\sigma_t^2$ as in Eq. \ref{eq:expected_variance}. This causes blurred results while applying averaging with DDPM, e.g., the second row of Figure \ref{fig:avg_wrong}. The reduced variance leads to over-homogeneous image content.

We propose the Variance-Corrected Fusion (VCF) by redefining $x^{(2)}_{t-1}$ to correct the variance:

\begin{equation}
\label{eq:post_avg_2}
\begin{aligned}
    x^{(2)}_{t-1} = &\sqrt{2} \frac{x^{(21)}_{t-1} + x^{(22)}_{t-1}}{2} + (1 - \sqrt{2}) \frac{\mu_{t}^{(21)} + \mu_{t}^{(22)}}{2} \\
    = &\frac{x^{(21)}_{t-1} + x^{(22)}_{t-1}}{\sqrt{2}} + (1 - \sqrt{2}) \frac{\mu_{t}^{(21)} + \mu_{t}^{(22)}}{2},
\end{aligned}
\end{equation}

so that have $x^{(2)}_{t-1} \sim N((\mu_{t}^{(21)} + \mu_{t}^{(22)}) / 2, \sigma_t^2)$.

We generalize the Eq. (\ref{eq:post_avg_2}) to averaging N overlaps:

\begin{equation}
    x_{t-1} = \frac{\sum_{i}^{N} x^{(i)}_{t-1}}{\sqrt{N}} + (1 - \sqrt{N}) \frac{\sum_i^N \mu_{t}^{(i)}}{N},
\end{equation}

and generalize to Guided Fusion weighted average:

\begin{equation}
\begin{aligned}
    x_{t-1} = &\frac{\sum_{i}^{N} w_i x^{(i)}_{t-1}}{\sqrt{\sum_i^N w_i^2}} \\
            & + (1 - \frac{W}{\sqrt{\sum_i^N w_i^2}}) \frac{\sum_i^N w_i \mu_{t}^{(i)}}{W},
\end{aligned}
\end{equation}

\noindent where $W = \sum_i^N w_i$.

The corrected formula can be applied to other SDE samplers that employ Gaussian noise, such as the EDM stochastic sampler \cite{karras_elucidating_2022}.

\subsection{One-shot Style Alignment (SA) for Coherent Montages}

SyncDiffusion \cite{lee_syncdiffusion_2023} inspires us that aligning the style of each small patch reduces the difficulty of generating more coherent content. However, SyncDiffusion requires constantly modifying the intermediate denoised patches to align their style, further disrupting the denoising process.

We noticed that the diffusion model exhibits the semantic interpolation effect \cite{song_denoising_2020}, in which the interpolations between two initial noises can lead to semantically meaningful results.

We propose a one-shot style-control method, Style Alignment (SA), performing interpolation on each non-overlapped patch cropped from the whole initial noise to a reference noise. The SA can be formulated as:

\begin{equation}
    \mathbf{x}^{(i)}_T := \text{slerp}(\mathbf{x}^{(i)}_T, \mathbf{z}^{\text{ref}}, \alpha)
\end{equation}

\noindent where the $\text{slerp}(\cdot)$ is the spherical linear interpolation \cite{shoemake_animating_1985} function; $\mathbf{x}^{(i)}_T$ is the $i^{th}$ non-overlapped crop from the initial noise $\mathbf{X}_T$; $\mathbf{z}^{\text{ref}}$ is a reference noise to be aligned with; $\alpha \in [0, 1]$ is the interpolation ratio where $0$ returns the original $\mathbf{x}^{(i)}_T$ and $1$ returns $\mathbf{z}^{\text{ref}}$. The reference noise $\mathbf{z}^{\text{ref}}$ can be any standard Gaussian noise. It may originate from a patch of the initial noise $\mathbf{X}_T$ or be obtained through diffusing a specific image.

After SA alignment, all non-overlapped patches rotate towards the reference noise, making them more clustered. Consequently, the distances between them are reduced, and their similarity increases.

\section{Results}

\textbf{Generated Datasets}. The text-to-panorama generation task was chosen to assess each method's performance on large-content image generation. For each approach, we sampled a set of $512 \times 3584$ sized images, $\times 7$ wider than the original model resolution, with ten prompts and 500 panorama images for each prompt. In total, 5,000 panorama images were generated for each approach. The panorama images were further divided into 7 patches matching the original model size, ultimately producing 35,000 images. The ten used prompts are:

\textit{1) A photo of a city skyline at night; 2) A photo of a mountain range at twilight; 3) A photo of a snowy mountain peak with skiers; 4) Cartoon panorama of spring summer beautiful nature; 5) Natural landscape in anime style illustration; 6) A photo of lush forest with a babbling brook; 7) An illustration of a beach in La La Land style; 8) Silhouette wallpaper of a dreamy scene with shooting stars; 9) A beach with palm trees; and 10) A film photo of a beachside street under the sunset.}

We conducted both qualitative and quantitative comparative experiments with the results obtained from MultiDiffusion (MD) \cite{bar-tal_multidiffusion_2023}, SyncDiffusion (Sync) \cite{lee_syncdiffusion_2023} and ElasticDiffusion (Elastic) \cite{elastic}.

\textbf{Reference Dataset}. Based on the prior works, the ODE samplers, such as DDIM, tend to lead to worse output quality \cite{song_denoising_2020, song_score-based_2020, karras_elucidating_2022}. We chose the SDE sampler DDPM to generate the reference dataset, which stands for higher quality. We used Stable Diffusion \cite{rombach_high-resolution_2022} v2.0 to generate reference images for evaluation. A reference dataset that contains 35,000 of $512 \times 512$ images was generated with 3500 images per prompt.

\textbf{Evaluation Metrics}. To assess the image quality, we employed FID \cite{heusel_gans_2017}, KID \cite{binkowski_demystifying_2018} (we use the anti-aliasing implementation \cite{parmar_aliased_2022}) and GIQA-QS/GIQA-DS \cite{vedaldi_giqa_2020} to evaluate the fidelity and diversity; CLIP score \cite{hessel_clipscore_2021} to evaluate the compatibility with the prompt.

\begin{figure}[t]
    \includegraphics[width=0.9\linewidth]{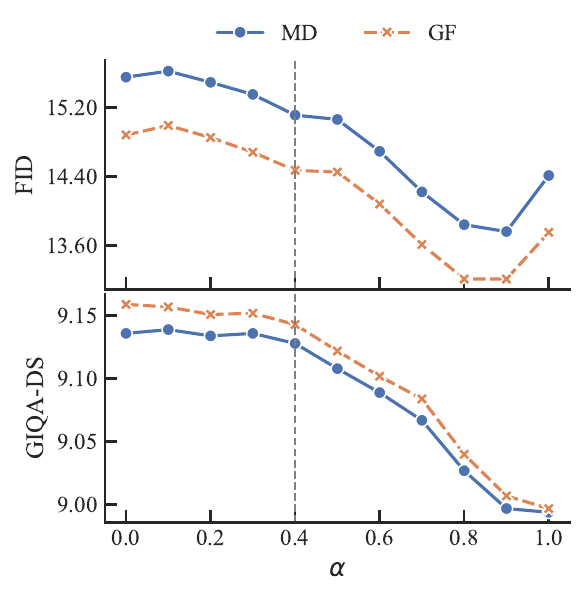}
    \caption{Image quality and diversity assessment using Style Alignment (SA) with different $\alpha$ values. The DDIM sampler is used.}
    \label{fig:sf_alpha}
\end{figure}

The results presented in Table \ref{tab:gf} and Figure \ref{fig:sf_alpha} were evaluated from the first five prompts. Table \ref{tab:overall} results were calculated from ten prompts.

\begin{figure}[ht]
    \centering
    \includegraphics[width=0.9\linewidth]{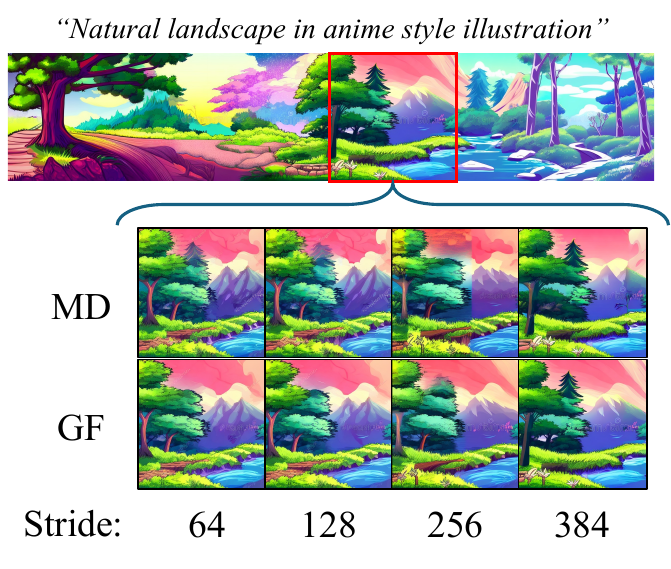}
    \caption{MultiDiffusion (MD) compared with Guided Fusion (GF) with different strides. All images are generated with the same initial noise.}
    \label{fig:gf}
\end{figure}

\begin{table}[h]
    \centering
    \begin{tabular}{c|c|c}
        Samplers & Methods  & Runtime (s) \\
        \hline\hline
        \multirow{4}{*}{DDIM} & MD      & 9.70±0.02   \\
        & Sync    & 102.31±0.52 \\
        & Elastic & 54.55±0.23  \\
        & GF      & 9.64±0.07   \\
        \hline
        DDPM & VCF & 181.16±0.31
    \end{tabular}
    \caption{Average runtime of single image generation. Results were calculated from 500 generations for each method.}
    \label{tab:runtime}
\end{table}

\subsection{The Effectiveness of Guided Fusion}

The stride controls the overlap ratio between patches; a smaller stride indicates a larger overlapping ratio. Additionally, a smaller stride indicates that more patches are needed in joint denoising to form a large image. Figure \ref{fig:gf} shows qualitative results from MultiDiffusion (MD) and Guided Fusion (GF) over 64, 128, 256, and 384 strides with a DDIM sampler. It can be observed that noticeable seams are present in the results of MD with four different strides. Among these, the seams are least apparent with the 64 stride, while they are most pronounced with the 256 stride. After applying GF, the seams are significantly reduced at all strides, resulting in more continuous images.

To thoroughly evaluate the effectiveness of the proposed GF, we compared our method with MD in three stride settings: 128, 256, and 384 with quantitative metrics. As shown in Table \ref{tab:gf}, the experimental results indicate that GF consistently outperforms MD across different strides. Specifically, GF achieved the best results in several key metrics, including FID, KID, GIQA-QS, and GIQA-DS, while MD demonstrated an advantage in CLIP scores. GF exhibited superior image quality and diversity, highlighting its greater applicability in fusing overlapped patches.

As shown in Table \ref{tab:gf}, as the stride increases, the FID and KID metrics of the results are gradually improved for both MD and GF. This supports our viewpoint: modifying the values in overlapping regions interferes with the denoising process of each patch and negatively affects the quality of the generated images. Although the seams are less evident with a higher overlap ratio, as the overlap ratio decreases, the FID and KID metrics of MD and GF gradually decrease, indicating that the generated images have better details.

We opted to use a stride of 384 for subsequent experiments because it demonstrated the best image quality and higher computational efficiency. Specifically, when generating a panorama image with a size of 512 $\times$ 3,584, employing a stride of 128 requires processing 25 patches, whereas using a stride of 384 requires only nine patches.

\subsection{High Image Quality Generation using DDPM Sampler with Variance-Corrected Fusion}

By examining Table \ref{tab:overall}, it can be observed that applying DDPM with VCF can produce high-quality and diverse outcomes. The "VCF" rows present substantial improvements to DDIM-based methods. We did not report the result from DDPM applied with MD because it produces blurred images, as shown in the second row of Figure \ref{fig:avg_wrong}. The third row of Figure \ref{fig:avg_wrong} and Figure \ref{fig:overall} shows that DDPM with VCF could generate large images with high contrast and fine details. The "VCF+GF" showing better scores than solely applying VCF indicates that the VCF and GF do not interfere with each other's effectiveness.

\begin{figure}[h]
    \centering
    \includegraphics[width=0.7\linewidth]{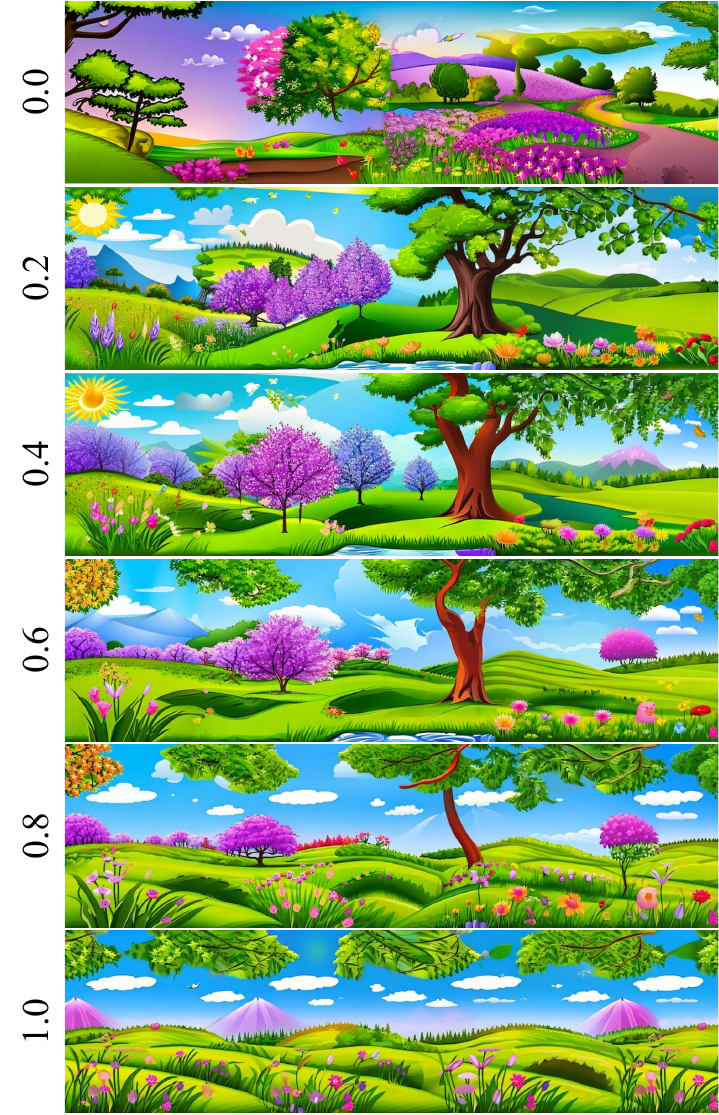}
    \caption{The left half of panorama images generated using Style Alignment (SA) with different $\alpha$ values. DDIM sampler is used.}
    \label{fig:sf_alpha_img}
\end{figure}

\subsection{The Effectiveness of Style Alignment}

For Style Alignment (SA), we use FID and GIQA-DS as the primary metrics to evaluate the quality and diversity of the generated panorama images. We assessed the generated images with $\alpha$ set to 0.0, 0.1, 0.2, ..., and 1.0 for MD and GF with the DDIM sampler. $\alpha=0.0$ implies the SA is not applied, while $\alpha=1.0$ initializes the entire large image using repeated reference noise patches.  We used a randomly generated standard Gaussian noise as the reference noise to conduct our experiments.

As shown in Figure \ref{fig:sf_alpha}, with the increase in $\alpha$, the overall image quality exhibits an upward trend, while diversity shows a downward trend. Figure \ref{fig:sf_alpha_img} shows progressive visual results from discontinuous content to the highly repeated pattern generated with increasing values of $\alpha$. This evidences our assumption: initializing patches with similarity helps to create more coherent content. The trade-off is that as $\alpha$ increases, diversity decreases. We identified the $\alpha=0.4$ as the optimal value because it balances the quality and diversity. With $\alpha$ larger than 0.4, the diversity drops quickly. The different choices of $\alpha$ provide a control of style consistency that can fit different aesthetic requirements.

It can also be observed from Figure \ref{fig:sf_alpha} that regardless of the choice of $\alpha$, applying SA with GF consistently achieves better quality and diversity compared to MD.

\begin{figure}[h]
    \centering
    \includegraphics[width=\linewidth]{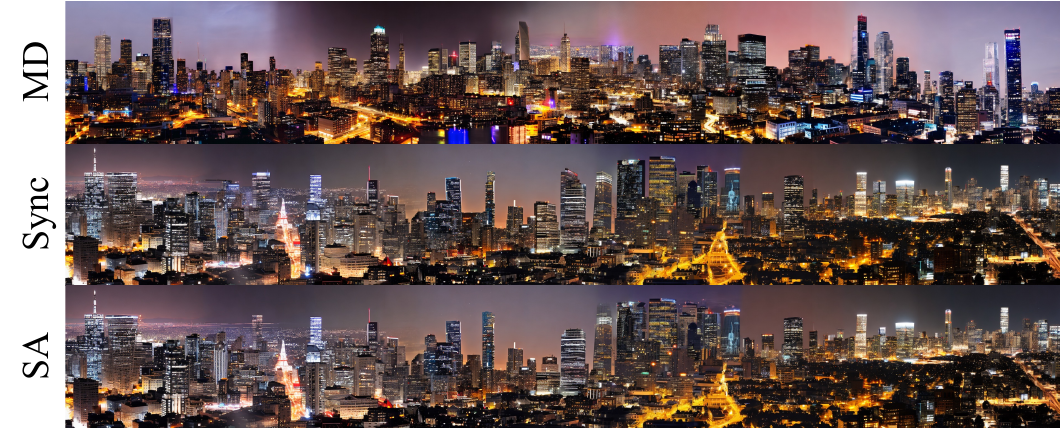}
    \caption{Highly similar results generated by SyncDiffusion (Sync) with a 0.1 sync weight and Style Alignment (SA) with $\alpha=0.1$. DDIM sampler is used to generate all results.}
    \label{fig:sync_sf_sim}
\end{figure}

As shown in Figure \ref{fig:sync_sf_sim}, we discovered that when using the same initial noise, the results generated by SyncDiffusion with a 0.1 sync weight and SA with $\alpha=0.1$ are highly similar to each other but significantly different from MD. In Table \ref{tab:sf_vs_sync}, we calculated the similarity between the images generated from three methods with DDIM sampler using Structural Similarity Index Measure (SSIM) \cite{wang_image_2004}, with 2500 panoramic images from each method. The SSIM between SA and SyncDiffusion reached 0.74, indicating that SyncDiffusion and SA produce highly similar outcomes. This implies that SA and SyncDiffusion are potentially equivalent to a specific content. Compared to SyncDiffusion, which uses gradient descent to align patch style at each denoising step, the SA is more computationally efficient as it only performs a one-shot alignment at initial noise. When generating a 3584-width image with the 384 stride, SA takes approximately 8s, while SyncDiffusion requires 102 seconds on a Quadro RTX 6000 card. The computational efficiency makes style control more feasible with the use of SDE samplers, which necessitates more denoising steps. The DDPM sampler requires 1000 denoising steps, which is 20 times longer than a 50-step DDIM sampler.

\begin{table}[h]
\centering
\begin{tabular}{c|c|c}
              & MD   & MD+SA$_{0.1}$ \\
\hline\hline
MD+SA$_{0.1}$ & 0.30 & --       \\
\hline
Sync          & 0.30 & 0.74     \\
\hline
\end{tabular}
\caption{SSIM Matrix.}
\label{tab:sf_vs_sync}
\end{table}

\section{Conclusions}

We have revisited joint denoising, which generates large images by creating a series of overlapped patches through small diffusion models. This work addresses the issues presented in the averaging fusion of overlapped regions, e.g., noticeable seams, blurred images, and discontinuous objects. We proposed a novel technique called Guided Fusion (GF), which reduces the disruption to the denoised image by assigning higher weights to the central region of each image patch, allowing the fused values in overlapped regions to be predominantly determined by the geometrically closer patch. Additionally, we presented Variance-Corrected Fusion (VCF), which adjusts the variance of the averaged values to enable its application with SDE samplers, such as DDPM. Furthermore, we introduced the Style Alignment (SA) that eases the fusion process by controlling the similarity of the initial noise, resulting in more coherent images.

Qualitative and quantitative experimental results demonstrate that the proposed methods effectively enhance the quality of the generated images. Our proposed approaches can be broadly applied to other joint denoising-based methods to achieve superior fusion outcomes. For instance, the high-resolution image generation approaches, ScaleCrafter \cite{he_scalecrafter_2024} and DemoFusion \cite{du_demofusion_2024}, both rely on MD to fuse the overlaps. Our approaches offer a potential enhancement for these approaches.

\bibliographystyle{ieeetr}
\bibliography{references}


\end{document}